%% file: main.tex
\documentclass[conference]{IEEEtran}
\pdfoutput=1
\IEEEoverridecommandlockouts
\usepackage{amsmath,amssymb,amsfonts}
\usepackage{graphicx}
\usepackage{textcomp}
\usepackage[numbers]{natbib} 
\usepackage{xcolor}
\def\BibTeX{{\rm B\kern-.05em{\sc i\kern-.025em b}\kern-.08em
    T\kern-.1667em\lower.7ex\hbox{E}\kern-.125emX}}
\usepackage{enumitem}
\setlist{nolistsep}
\usepackage{soul}
\usepackage{caption}
\usepackage[ruled,vlined]{algorithm2e}
\usepackage{algpseudocode}
\usepackage{graphicx}
\usepackage{subfig}
\usepackage{multirow}
\usepackage{wrapfig}
\usepackage{float}
\usepackage{booktabs} 

\newcommand{\QS}[1]{\textcolor{black}{#1}}

\newcommand{\rev}[1]{\textcolor{purple}{#1}}
\newcommand{\std}[2]{#1$\pm$#2}
\newcommand{\ie}{\emph{i.e.}}
\newcommand{\eg}{\emph{e.g.}}

\colorlet{purple}{black}

\newenvironment{colorsection}[1]
  {\color{#1}}
  {\par}

\input{math_commands}

\begin{document}

\title{Exploring Parameter-Efficient Fine-Tuning to Enable Foundation Models in Federated Learning
}

\author{\IEEEauthorblockN{1\textsuperscript{st} Guangyu Sun}
\IEEEauthorblockA{\textit{Center for Research in Computer Vision} \\
\textit{University of Central Florida}\\
Orlando, FL, USA \\
guangyu.sun@ucf.edu}
\and
\IEEEauthorblockN{2\textsuperscript{nd} Umar Khalid}
\IEEEauthorblockA{\textit{Center for Research in Computer Vision} \\
\textit{University of Central Florida}\\
Orlando, FL, USA \\
umar.khalid@ucf.edu}
\and
\IEEEauthorblockN{3\textsuperscript{rd} Matias Mendieta}
\IEEEauthorblockA{\textit{Center for Research in Computer Vision} \\
\textit{University of Central Florida}\\
Orlando, FL, USA \\
matias.mendieta@ucf.edu}
\and
\IEEEauthorblockN{4\textsuperscript{th} Pu Wang}
\IEEEauthorblockA{\textit{Department of Computer Science} \\
\textit{University of North Carolina at Charlotte}\\
Charlotte, NC, USA \\
pu.wang@uncc.edu}
\and
\IEEEauthorblockN{5\textsuperscript{th} Chen Chen}
\IEEEauthorblockA{\textit{Center for Research in Computer Vision} \\
\textit{University of Central Florida}\\
Orlando, FL, USA \\
chen.chen@crcv.ucf.edu}
}

\maketitle

\begin{abstract}
  Federated learning (FL) has emerged as a promising paradigm for enabling the collaborative training of models without centralized access to the raw data on local devices. In the typical FL paradigm (\eg, FedAvg), model weights are sent to and from the server each round to participating clients. 
Recently, the use of small pre-trained models has been shown to be effective in federated learning optimization and improving convergence.
However, recent state-of-the-art pre-trained models are getting more capable but also have more parameters, known as the ``Foundation Models."
In conventional FL, sharing the enormous model weights can quickly put a massive communication burden on the system, especially if more capable models are employed. 
Can we find a solution to enable those strong and readily available pre-trained models in FL to achieve excellent performance while simultaneously reducing the communication burden?
To this end, we investigate the use of parameter-efficient fine-tuning in federated learning and thus introduce a new framework: FedPEFT. Specifically, we systemically evaluate the performance of FedPEFT across a variety of client stability, data distribution, and differential privacy settings. By only locally tuning and globally sharing a small portion of the model weights, significant reductions in the total communication overhead can be achieved while maintaining competitive or even better performance in a wide range of federated learning scenarios, providing insight into a new paradigm for practical and effective federated systems.
\end{abstract}


\input{1_introduction}
\input{2_related_work}
\input{3_method}
\input{4_experiments}
\input{5_discussion}

\section{Acknowledgement}
This work is partially supported by the NSF/Intel Partnership on MLWiNS under Grant No. 2003198 and the NSF Grant No. 2008447.

\vspace{-2mm}
{\small
\bibliographystyle{unsrt}
\bibliography{main}
}

\end{document}

%% file: math_commands.tex
\pdfoutput=1
\usepackage{amsmath,amsfonts,bm}

\def\eqref#1{equation~\ref{#1}}

\def\1{\bm{1}}

\def\vtheta{{\bm{\theta}}}

\def\vh{{\bm{h}}}

\def\vx{{\bm{x}}}

\DeclareMathAlphabet{\mathsfit}{\encodingdefault}{\sfdefault}{m}{sl}
\SetMathAlphabet{\mathsfit}{bold}{\encodingdefault}{\sfdefault}{bx}{n}

\def\gO{{\mathcal{O}}}

\def\sD{{\mathbb{D}}}

\def\sS{{\mathbb{S}}}

\newcommand{\R}{\mathbb{R}}

\newtheorem{assumption}{Assumption}
\newtheorem{remark}{Remark} 
\newtheorem{theorem}{Theorem}


%% file: 1_introduction.tex
\pdfoutput=1
\section{Introduction}
Federated learning (FL)~\citep{mcmahan2017communication} has become increasingly prevalent in the research community, having the goal of enabling collaborative training with a network of clients without needing to share any private data.
One key challenge for this training paradigm is overcoming data heterogeneity.
The participating devices in a federated system are often deployed across a variety of users and environments, resulting in a non-IID data distribution. As the level of heterogeneity intensifies, optimization becomes increasingly difficult.
Various techniques have been proposed for alleviating this issue. These primarily consist of modifications to the local or global objectives through proximal terms, regularization, and improved aggregation operations~\citep{li_federated_2020, karimireddy2020scaffold, mendieta_local_2022, acar2021federated, wang2020tackling}.
More recently, some works have investigated the role of model initialization in mitigating such effects~\citep{nguyen_where_2022, chen_pre-training_2022}. Inspired by the common usage of pre-trained models for facilitating strong transfer learning in centralized training, researchers employed widely available pre-trained weights for initialization in FL and were able to close much of the gap between federated and centralized performance.

Still, while pre-trained initializations are effective for alleviating heterogeneity effects in FL, another key challenge is left unaddressed; that is, communication constraints. This is often the primary bottleneck for real-world federated systems \citep{kairouz_advances_2021}.
In the standard FL framework \citep{mcmahan_communication-efficient_2017}, 
updates for all model parameters are sent back and forth between the server and participating clients each round. This can quickly put a massive communication burden on the system, especially if more capable models beyond very small MLPs are used.

When employing strong pre-trained models, the number of parameters can be large, such as for current state-of-the-art transformers. For example, ViT-Base (ViT-B)~\citep{dosovitskiy_image_2021} has 84 million parameters, let alone the current significant progress in large foundation models (\eg, GPT-4~\citep{openai2023gpt4} has more than 1 trillion parameters). Those large models would simply exacerbate the communication overhead to insurmountable levels. \QS{As a compromise, most existing FL work focuses on the performance of smaller Convolutional Neural Networks (\eg, ResNet~\citep{he2016deep}) on smaller datasets (\eg, CIFAR-10~\citep{krizhevsky_learning_2012}, EMINIST~\citep{cohen2017emnist}). Considering the thriving progress in large pre-trained Foundation Models~\citep{bommasani_opportunities_2022}, \textit{an efficient framework enabling these large pre-trained models will be significant for the FL community.}} 

\begin{figure}[t]
\centering
\includegraphics[width=\linewidth]{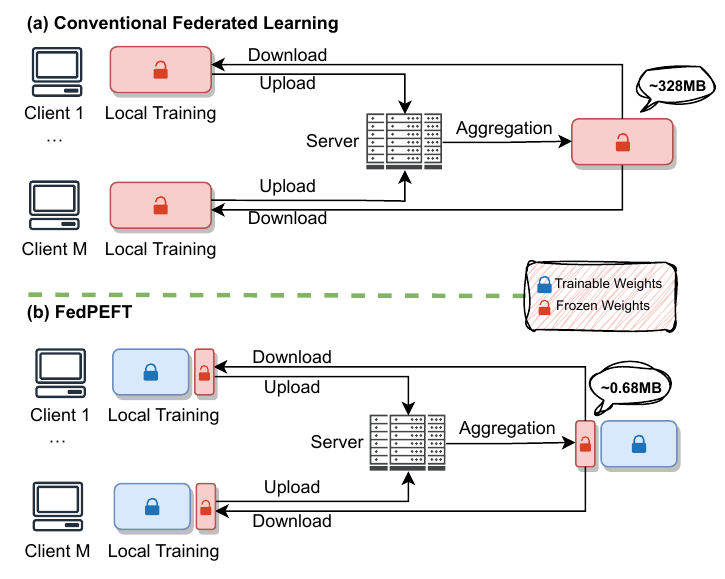}
\caption{\small\textbf{Process in a federated learning communication round with $\mathbf{M}$ participating clients.} We use ViT-Base as an instance to analyze the communication costs. (a) Conventional federated learning framework, where the entire model will be sent during the communication. (b) FedPEFT, which is our proposed parameter-efficient framework for federated learning. 
}
\label{fig:intro}
\vspace{-5mm}
\end{figure}
Based on the previous study on centralized training~\citep{cai_tinytl_2020, jia_visual_2022, pfeiffer_adapterhub_2020, chen_conv-adapter_2022}, we note that pre-trained models have strong representations, and updating all the weights during fine-tuning is often not necessary. Various parameter-efficient fine-tuning methods (\eg, fine-tuning only a subset of the parameters or the bias terms) for centralized training have been proposed in the literature and show that successful and efficient adaptation is possible, even under domain shift \citep{jia_visual_2022, cai_tinytl_2020, pfeiffer_adapterhub_2020}.
We reason that such insight is applicable to FL, where each client can be thought of as a shifted domain on which we are fine-tuning.
By leveraging pre-trained weights, it may be possible to simply update a small portion of the weights for each client. 
This will significantly reduce the communication burden on the system, as the updates communicated with the server will consist of just a fraction of the total model parameters.

Can we reap these potential communication benefits while still achieving strong performance in FL?
Unfortunately, operating conditions in FL are difficult, requiring successful convergence under varying data heterogeneity levels, random client availability, and differential privacy procedures. 
Therefore, we are unable to properly assess this possibility of benefit based on existing literature, as \textit{diverse parameter-efficient fine-tuning methods have not been systematically explored in such situations in FL.}
To fill this gap, we explore the viability of a Federated Parameter-Efficient Fine-Tuning (FedPEFT) framework \QS{with a systemic empirical study on a comprehensive set of FL scenarios} including \textit{communication analysis} about communication cost for each method to enable pre-trained models, \textit{capability analysis} of each method with unlimited communication budget, and \textit{robustness analysis} of each method when additional constraints (\ie, differential privacy or data scarcity) applied. The framework is illustrated in Fig.~\ref{fig:intro}. We deploy parameter-efficient fine-tuning methods to adapt pre-trained models and enable massive reductions in communication overheads.

The contribution of this paper is summarized as follows:

\setlist[itemize]{leftmargin=*}
\begin{itemize}
    \item We explored several PEFT methods in FL as the FedPEFT framework to simultaneously addresses data heterogeneity and communication challenges. 
    FedPEFT allows for the utilization of powerful pre-trained models in federated learning while keeping communication costs extremely low.
    \item We present a systematic study of the FedPEFT framework with various fine-tuning methods under heterogeneous data distributions, client availability ratios, and increasing degrees of domain gap relative to the pre-trained representations \QS{on both \textit{image} and \textit{video} domains}, showing the capability of FedPEFT. (Sections~\ref{sec:comm} and \ref{sec:cap})
    \item To ensure FedPEFT is practical for the complex environments of FL, we further analyze the robustness of FedPEFT among low-data regimes and differential privacy operations. (Sections~\ref{section:robust})
\end{itemize}

%% file: 2_related_work.tex
\pdfoutput=1
\section{Related Work}

\noindent\textbf{Federated Learning.} FL is a decentralized training paradigm composed of two procedures: local training and global aggregation. Therefore, most existing work focuses on either local training~\cite{mendieta_local_2022, li_model-contrastive_2021, li_federated_2020} or global aggregation~\cite{wang_tackling_2020, yurochkin_bayesian_2019} to learn a better global model. 
Another line of work cuts into this problem by applying different initializations to help both procedures. \cite{chen_pre-training_2022} shows that initializing the model with pre-trained weights can make the global aggregation of FedAvg more stable, even when pre-trained with synthetic data. Furthermore, \cite{nguyen_where_2022} presents the effectiveness of pre-training with different local and global operations. However, these works focus purely on the effect of initialization in a standard FedAvg framework and do not consider the communication constraints of the system. Our work pushes the envelope further by leveraging strong pre-trained models (even large, capable transformers) in FL while effectively handling the communication issue via parameter-efficient fine-tuning.

\noindent\textbf{Communication in Federated Learning.} Communication constraints are a primary bottleneck in federated learning. To reduce the communication cost, several previous work leverage model compression techniques~\cite{konecny_federated_2017, suresh_distributed_2017}. Such works do not change the training paradigm but rather post-process the local model to reduce communication costs. For instance, \cite{konecny_federated_2017} proposes approaches that parameterize the model with fewer variables and compress the model in an encoding-decoding fashion. However, the minimal requirement to maintain all the information is still high when facing today's large models. Meanwhile, another line of work changes the training paradigm by learning federated ensembles based on several pre-trained base models~\cite{hamer_fedboost_2020}. In this way, only the mixing weights of the base models will be communicated in each round. This approach aims to reduce the burden of downloading and uploading the entire model in each round.
However, the base models are not directly trained, and the final performance is highly related to the base models. Meanwhile, model ensembles will take more time and space, which is often limited on the client side. Our framework follows the strategy of this line of work that does not transmit the entire model, but we use only one pre-trained model instead of several base models and only transmit a subset of the parameters instead of the model ensembles. Therefore, no additional time or space is required.

\noindent\rev{\textbf{Parameter-Efficient Fine-tuning.} Fine-tuning is a prevalent topic in centralized transfer learning, especially in this era of the ``Foundation Model''~\cite{bommasani_opportunities_2022}. A significant line of work is to reduce the trainable parameter number, \ie, parameter-efficient fine-tuning (PEFT)~\cite{chen_conv-adapter_2022, pan_parameter-efficient_2022, liu_few-shot_2022, hu2021lora, liu_p-tuning_2022, zaken2021bitfit, cai_tinytl_2020, he2021towards}. PEFT has emerged as a pivotal area of research in the field of natural language processing (NLP)~\cite{pfeiffer_adapterhub_2020, pfeiffer_adapterfusion_2021,li2021prefixtuning,bahng_exploring_2022} and further adapted into more fields such as computer vision (CV)~\cite{jia_visual_2022, yang2023aim, yao_cpt_2022, chen2022adaptformer, jie2022convolutional, pan2022st, gao2023unified}. 
With different focuses, PEFT methods can be divided into three categories: 1) Input Adaptation methods such as prompt-tuning~\cite{jia_visual_2022, liu_p-tuning_2022} focusing on adding learnable context to the input data, 2) Backbone Adaptation methods such as adapter-tuning~\cite{chen2022adaptformer,pfeiffer_adapterhub_2020, pfeiffer_adapterfusion_2021}, and 3) Specification methods such as bias-tuning~\cite{cai_tinytl_2020, zaken2021bitfit}.
PEFT enables easier access and usage of pre-trained models by reducing the memory cost needed to conduct fine-tuning due to fewer computed gradients.
In federated learning, PEFT has an additional benefit that is not salient in centralized training: reducing communication costs. By introducing PEFT to federated learning, our work can take advantage of a strong (and even large) pre-trained model while significantly reducing communication costs. 
Several works study the PEFT methods under FL settings in NLP~\cite{zhang2023fedpetuning, zhao2023fedprompt, zhang_federated_2022, chen2022fedtune}. Complementarily, \textit{our work provides a comprehensive study of PEFT in various FL settings in computer vision under both image and video tasks and insights on various scenarios, including more privacy requirements or under limited data}. }

%% file: 3_method.tex
\pdfoutput=1

\section{Federated Parameter-Efficient Fine-Tuning}
\subsection{Problem Formulation}
In this section, we formally describe the federated learning objective and federated parameter-efficient fine-tuning. 
Using a classification task as an example, $K$ samples in a dataset $\sD = \{ (\vx_k,y_k)_{k=1}^K\}=\cup_{n=1}^{N} \sD_n$, where $\vx$ is the input and $y \in \{0,1,\dots,C-1\}$ is the label, are distributed among $N$ clients. Each client has a local model $\{\phi_i\}_{i=1}^N$ parameterized by $\{\vtheta_i \cup \delta_i\}_{i=1}^N$, where $\vtheta$ is the pre-trained weights and $\delta$ is the trainable parameters. 
The goal of federated learning is\textit{ to learn a global model} $\phi$ parameterized by $\vtheta\cup\delta$ on the server from $M$ sampled client models in $T$ communication rounds by minimizing the global objective $F$ as 
\begin{equation}
    \min_{\delta} F (\phi) = \frac{1}{|\sD_{test}|}\sum_{i=1}^{|\sD_{test}|} \ell (y_k, \phi_i^{(t)}(\vx_k))
\end{equation}
on a hold-off test set $\sD_{test}$ with a loss function $\ell$. \textit{Compared with traditional FL updating the entire $\phi$, only $\delta$ is updated in FedPEFT.}

At the beginning of training, $\vtheta^{(0)}$in
the global model $\phi^{(0)}$ is initialized with pre-trained weights, and $\delta^{(0)}$ is randomly initialized, where the superscript $t$ indicates the model at round $t$.
In each round $t$, $M$ clients will be selected for communication, and $\{\phi_i^{(t)}\}_{i=1}^M$ will be initialized by $\phi^{(t)}$ and updated by 
\begin{equation}
    \min_{\delta_i} F_i(\phi_i^{(t)})=\frac{1}{|\sD_i|} \sum_{k=1}^{|\sD_i|} \ell (y_k, \phi_i^{(t)}(\vx_k))
\end{equation}
for $E$ epochs,
where $F_i$ is the local objective. After the local updates, the server will receive and aggregate the trainable parameters $\{\delta_i^{(t)}\}_{i=1}^M$ with the FedAvg algorithm to a new global model 
\begin{equation}
 \phi^{(t+1)} = \sum_{m=1}^M \frac{|\sD_m|}{\sum_{i=1}^M|\sD_i|}\phi_i^{(t)}.
\end{equation}
This procedure is repeated from $t=0$ to $t=T-1$.
During the client-server communication, we only take the communication cost for the model into consideration, assuming the remaining communication costs are fixed. Therefore, the communication cost $C$ is proportional to the transmission parameters number, thus can be formulated as
\begin{equation}
    C \propto |\delta| \cdot M,
\end{equation}
\QS{We take the one-way communication cost (\ie, upload or download) as the metric.} \textit{The final goal of this problem is to minimize the $C$ while maintaining server accuracy.}



\subsection{FedPEFT}
\begin{figure}[t]
\vspace{-10pt}
\centering
\includegraphics[width=\linewidth]{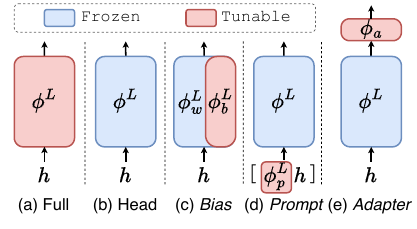}
\vspace{-10pt}
\caption{\small\rev{Methods to fine-tune each layer in a pre-trained backbone, where $h$ means the input, $\phi$ means the pre-trained layer, and $\phi_w, \phi_b$ mean its weight and bias parameters, respectively.
}}
\vspace{-15pt}
\label{fig:method}
\end{figure}

\begin{algorithm}[t]
\caption{Algorithm of FedPEFT framework}\label{alg:fedpeft}
\small

\begin{algorithmic}

\noindent \colorbox[rgb]{1, 0.95, 1}{
\begin{minipage}{0.9\linewidth}

\textbf{Input:} $N$ clients indexed by $i$, participating-client number $M$, communication rounds $T$, \textcolor{blue}{trainable parameters $\delta$ of the model where $|\delta| << |\phi| $}, pre-trained model weights $\vtheta$, random initialized $\delta^{(0)}$, and local epoch number $E$.

\end{minipage}
}

\colorbox[gray]{0.95}{
\begin{minipage}{0.9\linewidth}
\item  \textbf{Server executes:}
\item     \hspace*{\algorithmicindent} initialize $\phi^{(0)}$ and $\{\phi_i^{(0)}\}_{i=1}^N$ with $\vtheta$ and $\delta^{(0)}$
\item     \hspace*{\algorithmicindent} \textbf{for } each round $t=0,2,...,T-1$ \textbf{do}
\item     \hspace*{\algorithmicindent} \quad $\sS_t \leftarrow$ (random set of $M$ clients)
\item     \hspace*{\algorithmicindent} \quad \textbf{for} each client $i\in\sS_t$ \textbf{in parallel do}
\item     \hspace*{\algorithmicindent} \quad \quad $\delta_i^{(t)}\leftarrow \delta^{(t)}$  
\item     \hspace*{\algorithmicindent} \quad \quad $\delta_i^{(t+1)} \leftarrow \textbf{ClientUpdate}(\delta_i^{(t)}, i)$
\item     \hspace*{\algorithmicindent} \quad $ \delta^{(t+1)}= \sum_{i=1}^M \frac{|\sD_i|}{\sum_{i=1}^M|\sD_i|}\delta_i^{(t+1)}$ 
\item     \hspace*{\algorithmicindent} \quad $\phi^{(t+1)}\leftarrow \{\vtheta, \delta^{(t+1)}\}$
\item     \hspace*{\algorithmicindent} \textbf{return} $\phi^{(T)}$
\end{minipage}
}
\colorbox[rgb]{0.95, 0.98, 1}{
\begin{minipage}{0.9\linewidth}
\item  \textbf{ClientUpdate ($\delta, i$):}
\item     \hspace*{\algorithmicindent} $\delta \leftarrow$ perform local training on $\delta$ with $\sD_i$ for $E$ epochs
\item     \hspace*{\algorithmicindent} \textbf{return} $\delta$
\end{minipage}
}
\end{algorithmic}
\end{algorithm}
In conventional federated learning, updates for the entire model need to be repeatedly sent to and from the server, resulting in significant communication costs, especially when larger, more capable modern neural network architectures are employed.
To reduce this heavy burden, we deploy parameter-efficient fine-tuning methods to adapt pre-trained models to the local clients rather than fully fine-tuning all parameters, which is described in Algorithm~\ref{alg:fedpeft}.
In the FedPEFT framework, illustrated in Fig.~\ref{fig:intro}, only a small amount of parameters in the local model will be downloaded, trained, and uploaded in each communication round. For instance, FedPEFT reduces the size of communication each round from \textbf{328MB} (85.88M parameters)/Client to \textbf{0.68MB} (0.17M parameters)/Client  when using a pre-trained ViT-Base as the backbone in our default setting introduced in Section~\ref{section:setting}. 

To implement FedPEFT, we provide a canonical baseline approach (Head-tuning) and three prototypes leveraging different representative parameter-efficient fine-tuning methods (Bias, Adapter, and Prompt), which are detailed in the following.


To reduce the number of trainable parameters, one intuitive method, Head-tuning, is to freeze the backbone $\phi$ and only train the head $c$.
This method is historically the most common fine-tuning procedure, and therefore we use it as a baseline for FedPEFT. 
However, the adaptation ability of this method is limited, as no adjustment is made to the network representation prior to the final output head. This can be problematic in the presence of a more intense domain shift.
Therefore, we consider the following approaches as primary prototypes for FedPEFT:

\textbf{FedPEFT-Bias.} Bias-tuning~\cite{cai_tinytl_2020} aims to adapt the pre-trained backbone by only fine-tuning a specific group of parameters, the bias term. In this way, the backbone can be trained with moderate adjustments to prevent damaging the upstream representation. \rev{We show the output of each layer $\phi^{l}$ given hidden states $h$ as} 
$
    \vh := \mathcolor{blue}{\phi_w^L}\vh+\mathcolor{red}{\phi_b^L},
$
\rev{where $\phi_w^L$ and $\phi_b^L$ are weight and bias of $\phi_L$, blue parameters are frozen, while red parameters are trainable.}

\textbf{FedPEFT-Adapter.} Instead of directly tuning existing parameters in the backbone like Bias-tuning, Adapter-tuning~\cite{pfeiffer_adapterhub_2020} adds a few parameters called adapters inside the backbone $\phi$ instead. Usually, adapters will be deployed in each layer of the backbone to perform transformations on different levels of the pre-trained feature while the backbone stays frozen. \rev{We show the output of each layer $\phi^{l}$ given hidden states $h$ as} 
$
    \vh := \mathcolor{red}{\phi_a}(\mathcolor{blue}{\phi_w^L}(\vh)),
$\rev{where $\phi_a$ is the adapter.}

\textbf{FedPEFT-Prompt.} 
Prompt-tuning \cite{jia_visual_2022} takes a slightly different approach from the other fine-tuning methods. Specifically, it concatenates trainable parameters, called prompt embeddings, to the input embedding and hidden states in each layer. \rev{We show the output of each layer $\phi^{l}$ given hidden states $h$ as} 
    $\vh := \mathcolor{blue}{\phi_w^L}(\left[\mathcolor{red}{\phi_p^L}, \vh\right])$,
\rev{where $\phi_p^L$ is the prompt for layer $L$ and $\left[\cdot,\cdot\right]$ is the concat operation.}

We illustrate the differences between all baseline and prototype methods in Fig.~\ref{fig:method}. \rev{Besides the above prototypes, our framework is compatible with other PEFT methods such as LoRA~\cite{hu2021lora}.}

\subsection{\rev{Convergence Guarantee}}
\begin{colorsection}{purple}
Based on the convergence of FedAvg in \cite{mcmahan2017communication, karimireddy2020scaffold, ding2022delta}, in this section, we will comment on the convergence of FedPEFT. 
For ease of notation, we consider, at each round, $S$ to be the number of clients sampled. 
We require the following assumptions. 
\begin{assumption}\label{ass:minimum}
	\textit{(Global minimum)} 
	For the global objective $F$, there exists $\phi^\star$ such that, $F(\phi^\star)=F^\star\leq F(\phi)$, for all $\phi\in\R^d.$  
\end{assumption}

\begin{assumption}\label{ass:smoothness}\textit{($\beta$-Smoothness)} The loss function $f_{i}: \R^d\to \R$ at each node is $\beta$-smooth, i.e. $f_{i}(y)\leq f_{i}(x)+\nabla f_{i}(x)^\top(y-x)+\frac{\beta}{2}\|y-x\|^2$ for all $x,y\in\R^d$. 
\end{assumption}
\begin{remark}\label{remark:beta_smooth}
\textit{(PEFT-FT gap)} 
	The above assumption implies that $F$ is $\beta$-smooth. Therefore, the gap between PEFT and Full Fine-tuning can be proved as $|F(\vtheta^{(T)}\cup \delta^{(0)})-F(\vtheta^{(T)}\cup \delta^{(0)}) = \gO(\frac{\beta}{2}(\|\vtheta^{(T)}-\vtheta^{(0)}\|^2+\|\delta^{(T)}-\delta^{(0)}\|^2))|$.
\end{remark}

\begin{assumption}\label{ass:noise}
    There exist constants $G\ge 0, B\geq1$, such that for all $x\in \R^d$, the stochastic noise, $\xi_{i,t}$ follows  
	\begin{eqnarray*}
	\frac{1}{N}\sum_{i=1}^{N}\|\nabla f_{i}(x)\|^2\le G^2+B^2\|F(x)\|^2.
	\end{eqnarray*}
\end{assumption}

\begin{assumption}\label{ass:bounded_grad}\textit{(Bounded variance)}
  Let $g_{i}(\phi):= \nabla f_{i}(\phi,z_{i(k)})$ be the unbiased stochastic gradient of $f_{i}$ with bounded variance. That is, there exists, $\sigma\ge 0$ such that,  $\mathbb{E}_{z_{i(k)}}\left[\|g_{i}(\phi)-\nabla f_{i}(\phi)\|^2\right]\le \sigma^2,$ for all $\phi,i$, where $z_{i(k)}$ is the $k^{\rm th}$ sample data at the $i^{\rm th}$ client. 
\end{assumption}

Based on the vanilla FedAvg framework, we can give our main convergence result as 
\begin{theorem}
   Let $F$ satisfies Assumptions \ref{ass:minimum}-\ref{ass:bounded_grad}. Then 
   $$
   \mathbb{E}\left[ \|\nabla F( \phi^{(T)})\|^2 \right] \le \gO\left(\frac{\beta\sqrt{(F(\phi^{(0)})-F^\star)}}{\sqrt{TEM}} + P\right),
   $$ where $P = \frac{\beta}{2}(\|\vtheta^{(T)}-\vtheta^{(0)}\|^2+\|\delta^{(T)}-\delta^{(0)}\|^2)$.
\end{theorem}
\end{colorsection}

\subsection{\rev{Privacy Discussion}}
\rev{Federated learning inherently ensures data privacy, as it keeps the training data localized. Consequently, our proposed FedPEFT framework, by design, does not introduce any additional risk of privacy leakage beyond what is intrinsic to FL itself. However, it is crucial to acknowledge the vulnerability of FL systems to gradient inversion attacks~\cite{huang_evaluating_2021, hatamizadeh_gradvit_2022}, where an adversary could potentially reconstruct original training data from shared gradients. This type of attack typically requires smaller batch sizes to be effective, as the precision of the information contained within the gradients diminishes with larger batch sizes, significantly reducing the feasibility of such attacks~\cite{dimitrov2024spear}. In light of this, FedPEFT inherently encourages the use of larger batch sizes compared to conventional full fine-tuning methods, as FedPEFT requires gradients for much fewer parameters and, therefore, very little memory cost during training. Enabling larger batch sizes not only optimizes the training efficiency but also fortifies the privacy-preserving nature of the federated learning framework, further mitigating the risks associated with gradient inversion attacks.}

%% file: 4_experiments.tex
\pdfoutput=1
\begin{table*}[t]
\caption{\small \textbf{Communication analysis on the image (upper) and video (lower) domains.} The communication cost is computed with 4B/parameter. The averaged final accuracy, \ie, $t=T=50$, and the standard deviation of three different random seeds are reported for each data set. The number of tuned parameters is computed based on CIFAR-100, but it may be slightly different for each dataset.
The first section shows the change in accuracy when decreasing the participating client number. The gray numbers indicate the baseline performance with no decrease in the participating client number. The second section shows the change in accuracy when we reward the low communication cost of head-tuning by increasing the number of participating clients. The third section shows the accuracy when we fully fine-tune a lightweight model, ShuffleNet V2 $0.5\times$~\cite{zhang_shufflenet_2017} for image domain or X3D-S~\cite{feichtenhofer2020x3d} for video domain. The fourth section shows the performance of each prototype of FedPEFT.}
\label{tab:main}
\small
\begin{center}
\resizebox{0.75\linewidth}{!}{
\begin{tabular}{cccc|ccc}
\toprule
\bf Model& \bf Method & \bf \# Tuned Params $\times$ \bf Clients & \bf Comm. Cost & \bf Resisc45 & \bf CIFAR-100 & \bf PCam\\
\midrule
ViT-B & Full Fine-tuning& 85.88M $\times$ 8 &2.56GB&\textcolor{gray}{91.49$\pm$0.82}& \textcolor{gray}{\std{91.73}{0.43}}&\textcolor{gray}{\std{85.41}{2.41}}\\
ViT-B & Full Fine-tuning& 85.88M $\times$ 4 &1.28GB&\textbf{92.13$\pm$0.87}& \std{89.69}{0.30}&\std{81.93}{3.54}\\
ViT-B & Full Fine-tuning& 85.88M $\times$ 2 &656MB&87.68$\pm$1.32& \std{87.03}{0.18}&\std{82.20}{1.22}\\
ViT-B & Full Fine-tuning& 85.88M $\times$ 1 &328MB&73.38$\pm$1.95& \std{74.79}{0.77}&\std{80.18}{1.83}\\
\midrule
ViT-B & Head-tuning & 0.08M $\times$ 8 &2.44MB&77.30$\pm$1.03&\std{72.45}{0.08}&\std{74.82}{2.40}\\
ViT-B & Head-tuning & 0.08M $\times$ 64 &19.53MB&83.58$\pm$0.45&\std{75.45}{0.16}&\std{77.82}{0.37}\\
\midrule
ShuffleNet& Full Fine-tuning& 0.44M $\times$ 8&13.43MB   &63.52$\pm$0.50&\std{49.81}{1.94}&\std{76.52}{3.35}\\
\midrule
ViT-B & FedPEFT-Bias & 0.18M $\times$ 8& 5.49MB &\textbf{89.04$\pm$0.80}&\textbf{\std{90.79}{0.25}}&\std{85.51}{0.66}\\
ViT-B & FedPEFT-Adapter & 0.23M$\times$ 8& 7.02MB&87.20$\pm$0.78&\std{87.74}{0.55}&\std{78.67}{1.85}\\
ViT-B & FedPEFT-Prompt&  0.17M$\times$ 8&5.19MB &83.35$\pm$0.76&\std{89.78}{0.84}&\textbf{\std{86.50}{0.85}}\\
\bottomrule
\toprule
\bf Model& \bf Method & \bf \# Tuned Params $\times$ \bf Clients & \bf Comm. Cost & \bf UCF101 & \bf HMDB51 & \bf UCF-CRIME\\
\midrule
ViT-B & Full Fine-tuning& 86.30M $\times$ 4 &1.29GB&\textcolor{gray}{94.22$\pm${0.23}}&\textcolor{gray}{70.34$\pm${0.28}}& \textcolor{gray}{34.37$\pm${0.76}}\\
ViT-B & Full Fine-tuning& 86.30M $\times$ 2 &656MB&93.85$\pm${0.33}&69.06$\pm${0.48}&32.03$\pm${0.26}\\
ViT-B & Full Fine-tuning& 86.30M $\times$ 1 &328MB&92.61$\pm${0.16}&59.88$\pm${0.27}&24.21$\pm${0.26}\\
\midrule
ViT-B & Head-tuning & 0.08M $\times$ 4 &1.22MB&88.57$\pm${0.30}&64.98$\pm${0.32}&32.03$\pm${0.76}\\
ViT-B & Head-tuning & 0.08M $\times$ 32 &9.76MB&89.74$\pm${0.45}&63.87$\pm${0.66}&32.81$\pm${0.76}\\
\midrule
X3D-S& Full Fine-tuning& 3.07M $\times$ 4&47.96MB&36.68$\pm${1.67}&27.74$\pm${0.24}&21.42$\pm${0.76} \\
\midrule
ViT-B & FedPEFT-Bias & 0.18M $\times$ 4&2.75MB&92.34$\pm${0.30}&69.37$\pm${0.32}&34.38$\pm${0.52} \\
ViT-B & FedPEFT-Adapter & 0.23M$\times$ 4&3.51MB&92.82$\pm${0.44}&69.96$\pm${0.24}&33.76$\pm${0.26}\\
ViT-B & FedPEFT-Prompt&  0.17M$\times$ 4&2.60MB&\textbf{\std{93.82}{0.66}}&\textbf{\std{70.87}{0.16}}& \textbf{\std{34.38}{0.26}}          \\
\bottomrule
\end{tabular}
}
\vspace{-15pt}
\end{center}
\end{table*}

\section{Experiments} \label{sec:experiments}
\QS{The capability of full fine-tuning in terms of accuracy has been illustrated in recent work~\cite{nguyen_where_2022, chen_pre-training_2022}. Thus we regard it as a competitive baseline for FedPEFT. }
To verify the performance of FedPEFT comprehensively, we evaluate the server accuracy with each method from three perspectives and aim to answer the following questions:


 \textbf{Communication Analysis:} When faced with a limited communication budget, there are several solutions to reduce costs, \eg, sampling fewer clients each round or using a lightweight model. Can FedPEFT outperform other solutions in terms of \textit{communication cost and accuracy}? \textbf{(RQ1)}

 \textbf{Capability Analysis:} When the communication budget is amply sufficient for all approaches, we want to evaluate the trade-off of training fewer parameters with FedPEFT. Can FedPEFT outperform full fine-tuning and training from scratch within \textit{various federated learning settings} and \textit{increasing levels of downstream domain gap}? \textbf{(RQ2)}

 \textbf{Robustness Analysis:} In a lot of application scenarios, there will be additional challenges for FL, such as privacy-preserving requirements (\ie, differential privacy) and data scarcity (\ie, very small amount of data on each client). We want to evaluate the robustness of each method under such scenarios. Can FedPEFT outperform full fine-tuning in terms of \textit{robustness}? \textbf{(RQ3)}


\subsection{Experiments details}\label{section:setting}
\textbf{Dataset.} For our study, we focus on computer vision (CV) applications as our testbed. Specifically, we investigate the performance of each method on the \textit{Image} and \textit{Video} domains with image classification and action recognition tasks. 
For the image classification, we employ \textbf{ImageNet-21K}~\cite{ridnik_imagenet-21k_2021} as the pre-training dataset. Then we select three datasets for the downstream tasks that have \textit{increasing degrees of domain gap} compared to ImageNet-21k, and we visualize and quantify the domain gap in Section~\ref{section:domain}: \QS{Resisc45~\cite{cheng2017remote}, } CIFAR-100~\cite{krizhevsky_learning_2012} and PCam~\cite{veeling_rotation_2018}. 
For video domain analysis, we take video action recognition task for evaluation. we employ \textbf{Kinetics-400}~\cite{kay2017kinetics} as the pre-training dataset and select three datasets with varying degrees of domain gap as compared to Kinetics-400: UCF101~\cite{soomro2012ucf101}, HMDB51~\cite{kuehne2011hmdb} and UCF-CRIME~\cite{sultani2018real}. 

\noindent\textbf{Experimental Setting.} Our default experimental setting is to split the dataset across $N=64$ clients and sample $M=8$ clients each round. The global aggregation will be performed after $E=10$ local epochs. A total of $T=50$ rounds of communication will be performed. To simulate heterogeneous data, we partition samples in each class to all clients following a Dirichlet distribution, as common in the literature \cite{mendieta_local_2022, acar2021federated, li_model-contrastive_2021}, with $\alpha=0.1$ for CIFAR-100 \QS{and Resisc45} and $\alpha=0.5$ for PCam based on the class number. Any modifications to this setting in subsequent experiments will be clearly noted.
 
 For each action recognition experiment, the data is split across $N=32$ clients and sample $M=4$ clients each round with constant $\alpha=0.1$ across all three video datasets. 

\noindent\textbf{Implementation Detail.} We choose ViT-B~\cite{dosovitskiy_image_2021} with image size 224 and patch size 16 as our backbone. For the image domain,
the backbone is pre-trained on ImageNet-21K~\cite{ridnik_imagenet-21k_2021}, as available in the timm library~\cite{wightman_pytorch_2019}.
The images for the downstream datasets are resized to $224 \times 224$. Images from CIFAR-100 are augmented by random cropping with a padding of $4$ and random horizontal flipping, and Resisc45 and PCam are augmented only with random horizontal flipping. For the video domain, we choose vanilla ViT-Base (ViT-B/16)~\cite{dosovitskiy_image_2021} with joint space-time attention as our backbone model~\cite{bertasius2021space} using VideoMAE~\cite{tong2022videomae} pre-trained weights on Kinetics-400~\cite{kay2017kinetics}. We perform the experiments on 8 Nvidia RTX A5000 GPUs with a batch size of $64$. All reported main results are run under $3$ random seeds and averaged. 



\subsection{RQ1: Communication Analysis}
\label{sec:comm}

To verify the effectiveness of FedPEFT and answer the first research question (\textbf{RQ1}, Section~\ref{sec:experiments}), we compare it with three baselines while monitoring the communication budget: a) Full fine-tuning of our default model (ViT-B). We vary the number of participating clients to show different levels of communication requirements. b) Head-tuning. The communication cost of head-tuning is naturally lower than other methods, so we increase the participating clients to make it a stronger baseline.
c) Fully fine-tune a light-weighted model (ShuffleNet V2 0.5$\times$ \cite{zhang_shufflenet_2017} for images and X3D-S~\cite{feichtenhofer2020x3d} for videos) with a similar communication overhead. 

\begin{figure}[t]
\vspace{-8pt}
\centering
\includegraphics[width=\linewidth]{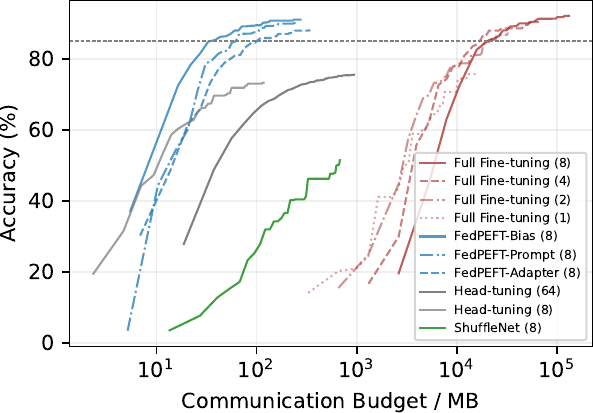}
\caption{\small\textbf{Server accuracy given the total communication budget.} The communication cost is computed with 4B/parameter, and the max number of communication rounds is 50. The number in the bracket next to the method indicates the number of participating clients $m$. The transparency of the line indicates the ratio between $m$ and total client number $N=64$. The horizontal dashed line shows a target accuracy of $85\%$.} 
\vspace{-15pt}
\label{fig:comm}
\end{figure}
As demonstrated in Table~\ref{tab:main}, all FedPEFT methods achieve better results in many cases compared with other approaches, even with significantly fewer communicated parameters. 
We find that full fine-tuning needs several orders of magnitude of communication to achieve a comparable result with FedPEFT. For instance, it needs
at least \textbf{187$\times$} and \textbf{477$\times$} more parameters to reach and outperform FedPEFT on CIFAR-100. Interestingly, full fine-tuning performs well on Resisc45 where the domain gap is smaller, even when the participating-client number is low. However, when the domain gap increases, more participating clients will be needed to outperform FedPEFT, and finally it fails to outperform FedPEFT even without reducing the participating-client number on PCam where a large domain gap exists. Meanwhile, 
head-tuning lags behind most other approaches, but the performance is stable with different levels of domain gap, while the ShuffleNet model only achieves $71\%$, $55\%$, and $89\%$ of accuracy on Resisc45, CIFAR-100, and PCam with \textbf{2.4$\times$} the communication cost compared with FedPEFT-Bias. Besides, the standard deviation of FedPEFT is lower than most other solutions, especially when the domain gap is large, showing the stability of FedPEFT. For the video domain, the conclusion is consistent.


In Fig.~\ref{fig:comm}, we also report the server accuracy that can be achieved for each method given the communication budgets using CIFAR-100 as an example. 
The communication cost per communication round for full fine-tuning is even higher than the total communication cost for FedPEFT to converge to similar final server accuracy. Meanwhile, all FedPEFT prototypes only require megabytes level communication, while full
fine-tuning requires gigabytes level communication to reach a given target accuracy (\eg, 85\% in Fig.~\ref{fig:comm}), showing the efficiency of FedPEFT.
For the inter-prototype comparison, FedPEFT-Bias stands out for its highest efficiency. We provide further discussions on the performance of each prototype in Section~\ref{sec:cap}.

\begin{table*}[t]
\vspace{-5pt}
\caption{\small\textbf{Capability analysis for different federated learning settings on CIFAR-100 and UCF-101.} 
Training from scratch in more complicated settings will lead to a lower result than in the first setting, which is omitted here.
Bold-style shows the best performance among all methods or among prototypes in FedFEFT.}
\vspace{-5pt}
\label{tab:all}
\small
\begin{center}
\resizebox{0.85\linewidth}{!}{
\begin{tabular}{cl|ccc|ccc}
\toprule
\multirow{2}{*}{\bf Client }& \multirow{2}{*}{\bf Method} & \multicolumn{3}{c|}{\bf Image} & \multicolumn{3}{c}{\bf Video} \\

&& \# Tuned Params&  Homogeneous &  Heterogeneous & \# Tuned Params&  Homogeneous &  Heterogeneous \\
\midrule
\multirow{6}{*}{$\begin{aligned}&N=16\\&M=16\end{aligned}$} & Scratch&85.88M & 38.44 & 35.72&86.30M&27.34&22.57 \\
 & Full Fine-tuning&85.88M & \textbf{93.70} & \textbf{93.50}&86.30M&\textbf{95.32}&\textbf{95.17} \\
 & Head-tuning&0.08M & 78.11 & 77.59&0.08M&89.90&85.44 \\
\cmidrule{2-8}
 & FedPEFT-Bias&0.18M & 91.89 & 90.25&0.18M&93.14&92.88 \\
 & FedPEFT-Adapter&0.23M & 90.21 & 88.77&0.23M&93.73&93.48 \\
 & FedPEFT-Prompt&0.17M & \textbf{92.09} & \textbf{90.37} &0.17M&\textbf{94.23}&\textbf{94.09} \\
\toprule
\multirow{5}{*}{$\begin{aligned}&N=16\\&M=2\end{aligned}$} & Full Fine-tuning&85.88M & \textbf{93.32} & \textbf{87.01} &86.30M&\textbf{95.12}&\textbf{94.78}\\
 & Head-tuning&0.08M & 76.65 & 62.80&0.08M&89.68&84.32 \\
\cmidrule{2-8}
 & FedPEFT-Bias&0.18M & 91.35 & 86.18&0.18M&93.05&92.64 \\
 & FedPEFT-Adapter&0.23M & 89.48 & 80.08&0.23M&93.53&93.27 \\
 & FedPEFT-Prompt&0.17M & \textbf{91.60} & \textbf{\textbf{85.54}}&0.17M&\textbf{94.11}&\textbf{93.89} \\
\toprule
\multirow{5}{*}{$\begin{aligned}&N=64\\&M=64\end{aligned}$} & Full Fine-tuning&85.88M & \textbf{93.66} & \textbf{92.81}&86.30M&\textbf{79.21}&\textbf{78.78} \\
 & Head-tuning&0.08M & 78.45 & 75.51&0.08M&59.56&52.74 \\
\cmidrule{2-8}
 & FedPEFT-Bias&0.18M & \textbf{92.71} &\textbf{ 91.71}&0.18M&78.21&75.33 \\
 & FedPEFT-Adapter&0.23M & 90.50 & 89.26&0.23M&78.41&75.88 \\
 & FedPEFT-Prompt&0.17M & 91.87 & 90.96&0.17M&\textbf{78.69}&\textbf{76.11} \\
\toprule
\multirow{5}{*}{$\begin{aligned}&N=64\\&M=8\end{aligned}$} & Full Fine-tuning&85.88M & \textbf{93.50} & \textbf{92.09} &86.30M&\textbf{78.78}&\textbf{76.66} \\
 & Head-tuning&0.08M & 77.59 & 72.55&0.08M&56.54&46.62 \\
\cmidrule{2-8}
 & FedPEFT-Bias&0.18M & \textbf{92.49} & \textbf{91.02}&0.18M&77.19&74.42 \\
 & FedPEFT-Adapter&0.23M & 90.39 & 88.05&0.23M&77.69&75.18 \\
 & FedPEFT-Prompt&0.17M & 92.00 & 89.90 &0.17M&\textbf{78.56}&\textbf{75.38} \\
\bottomrule
\end{tabular}
}
\end{center}
\vspace{-13pt}
\end{table*}





\subsection{RQ2: Capability Analysis}
\label{sec:cap}
To study and understand our second research question (\textbf{RQ2}, Section~\ref{sec:experiments}), we analyze the impact of the domain gap between the model pre-training dataset and the dataset for FL (Section~\ref{section:domain}) and systemically perform experiments on CIFAR-100 across different federated learning scenarios by varying client status and data distribution (Section~\ref{section:diff_FL_setting}).
\begin{figure}[t]
\centering
\begin{minipage}{0.45\linewidth}
\subfloat[Extracted feature. The number in the bracket is the domain gap to the pre-training dataset.]{\includegraphics[width=\linewidth]{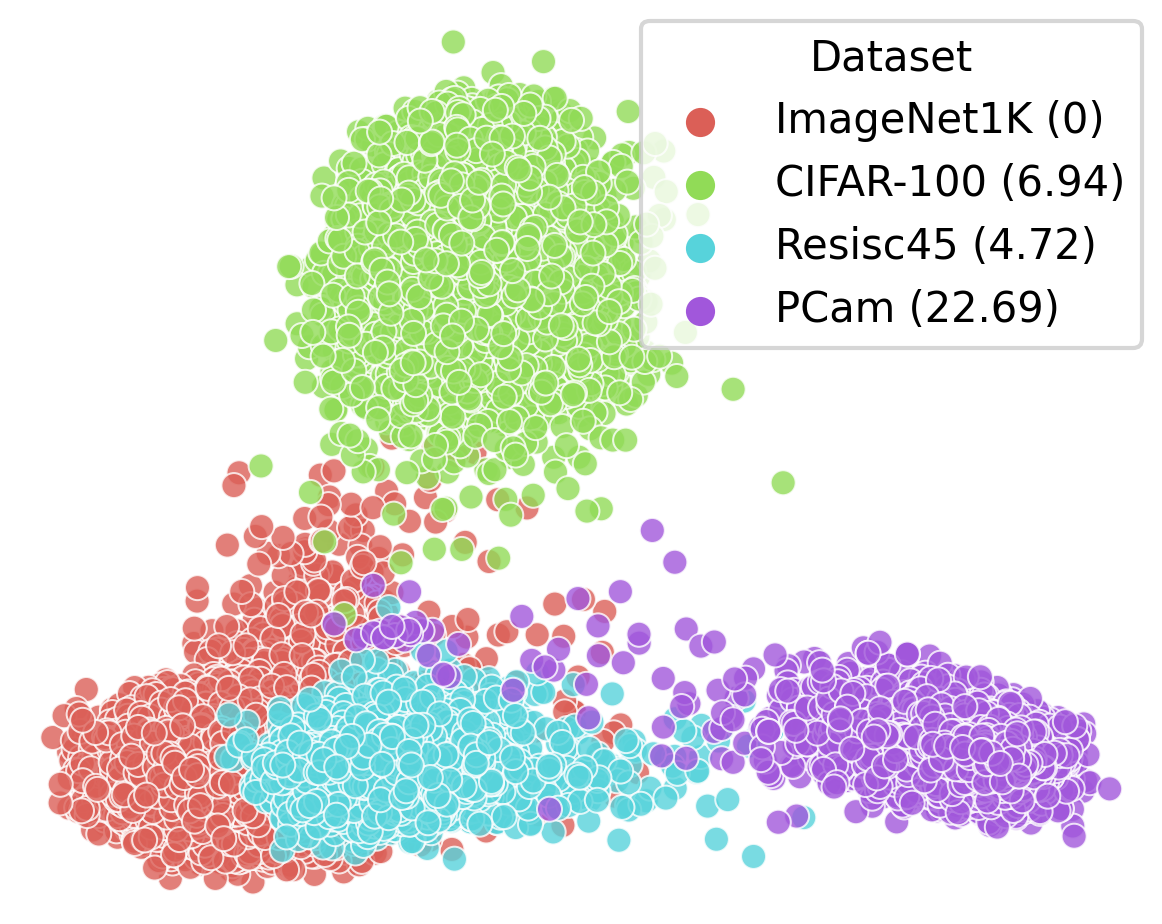}
        \label{fig:lda}}
\end{minipage}
\hfill
\begin{minipage}{0.45\linewidth}
\subfloat[Relative accuracy of each method compared with full fine-tuning with different domain gaps.]
{\includegraphics[width=\linewidth]{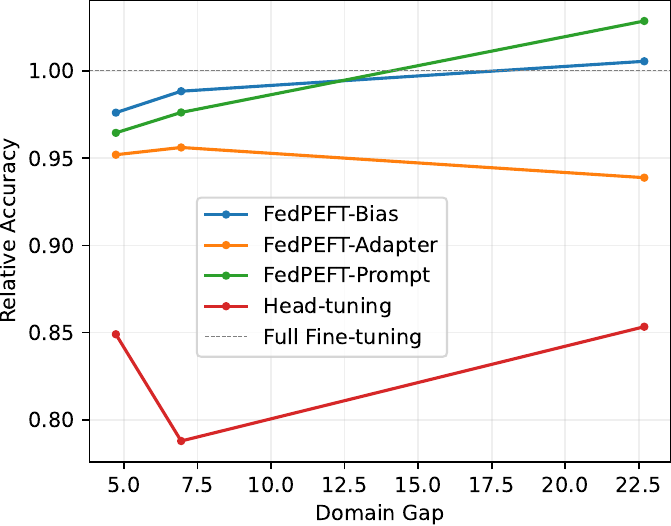}
        \label{fig:dva}}
\end{minipage}
\caption{\small\textbf{Visualization and analysis of domain gap.} 
}
\vspace{-5mm}
\end{figure}
\noindent\textbf{Capability with Domain Gap.} \label{section:domain}
\QS{Domain gap is a realistic concern when deploying pre-trained models for downstream tasks. As shown in Tab.~\ref{tab:main}, the performance on different datasets varies a lot. Besides the difference in the difficulty between all datasets, the domain gap to the pre-trained dataset is also a key concern here since Resisc45 is a remote sensing dataset, and PCam is a medical dataset. To further discuss the impact of the domain gap for each method, we use our default setting shown in Section~\ref{section:setting} on the image domain as an instance to visualize and quantify the domain gap between each downstream dataset and the pre-training dataset. Specifically, we adapt Linear Discriminant Analysis (LDA) for all extracted features for the test samples in each dataset from the pre-trained backbone to reduce the dimension. We compute the center of each dataset and then compute the \textbf{distance to the center} of the pre-training dataset as the quantifying result of the domain gap, as shown in Fig.~\ref{fig:lda}.}

In Fig.~\ref{fig:dva}, we present the performance of all approaches with an increasing degree of domain gap compared to the ImageNet-21k pre-training dataset. Interestingly, full fine-tuning falls further behind as the data domain gap widens in the PCam scenario, largely unable to keep up with FedPEFT despite requiring a massive communication budget. This phenomenon when a pre-trained model meets out-of-domain data has been studied under centralized settings~\cite{shen_partial_2021}. It was found that the pre-trained upstream representations are still meaningful even with a domain gap. Therefore, fully fine-tuning the backbone with out-of-domain data can damage the high-level semantics inside these upstream representations due to overfitting, especially when the data size is small. This is particularly relevant in FL, where overfitting and subsequent client drift \cite{varno_minimizing_2022, mendieta_local_2022,  karimireddy2020scaffold} are prone to occur.

On the opposite end of the spectrum, by not fine-tuning the backbone at all, head-tuning maintains similar accuracy despite the domain gap. 
This shows the robustness of the pre-trained high-level semantics across domains, supporting the conclusion that \textbf{there is meaningful high-level semantics inside of the upstream representations}.

Still, the tight restriction on head-tuning is perhaps a bit too far, as the accuracy on all datasets is still low overall. Between the two extremes of head-tuning and full fine-tuning, FedPEFT approaches may be able to suitably adapt the upstream representations without excessively damaging them.
Specifically, FedPEFT-Bias operates with parameter-level control for each parameter pair containing weight and bias terms. The representation can then preserve the high-level semantics by freezing the weight term (maintaining the direction in the feature space) and still adapting via the bias term (shifting in the feature space). FedPEFT-Adapter and FedPEFT-Prompt have slightly different mechanisms, controlling the backbone by transforming the intermediate hidden representations via adapters and prompts. 
\rev{Specifically, FedPEFT-Prompt adds additional hidden states before the original hidden states without changing the original representation, while FedPEFT-Adapter transforms hidden states into a new space. Consequently, FedPEFT-Prompt shows stronger robustness in handling larger domain gaps than FedPEFT-Adapter.}
Of these approaches, FedPEFT-Prompt is the most stable under the domain gap, surpassing full fine-tuning by ~1.1\% on PCam.
Overall, we hypothesize that \textbf{more fine-tuning freedom will be better when the domain gap is minor, but moderate fine-tuning is needed to maintain, as well as control, the high-level semantics when the domain gap is large}.
\noindent\textbf{Capability with Different FL Settings.}\label{section:diff_FL_setting}
In application scenarios, the setting of federated learning can vary substantially. It is important to show the capability to maintain high performance in diverse settings. We present results for all approaches with different client availability ratios and data distributions in Table~\ref{tab:all} and draw the following conclusions from the experiments: 

First, we see that \textbf{fine-tuning the pre-trained model shows a significant improvement over training from scratch}, especially in heterogeneous scenarios. This finding is in agreement with other very recent works~\cite{chen_pre-training_2022, nguyen_where_2022}, which note the stabilization effect of pre-trained initialization in federated optimization.
When only fine-tuning the head, the performance is still much better than training the entire model from scratch but remains low in comparison to other methods across all settings. We again find that head-tuning simply lacks adaptation ability, holding too closely to the upstream representation.


On the other hand, we find that \textbf{FedPEFT achieves comparable results \rev{($\mathbf{\ge 95\%}$)} to full fine-tuning with less than 0.3\% of the trainable parameters.} This ability to maintain accuracy performance in various scenarios is crucial for FL, as oftentimes, the exact setting and distributions are not known ahead of time. 
\rev{Meanwhile, for inter-prototype comparison, we find that \textbf{FedPEFT-Bias outperforms other prototypes in almost all settings in the image domain, while FedPEFT-Prompt shows leading performance among all prototypes in the video domain.} This provides guidance in application to choose the prototype.}


\vspace{-3mm}
\subsection{RQ3: Robustness Analysis}\label{section:robust}

\begin{table}[t]
\vspace{-4mm}
\caption{\small\textbf{Robustness analysis for privacy-preserving.}  The red number indicates the performance difference.}
\label{tab:dp}
\begin{center}
\vspace{-2mm}
\resizebox{\linewidth}{!}{%
\begin{tabular}{l|cc| cc}
\toprule
\bf \multirow{2}{*}{Method} & \multicolumn{2}{c|}{\bf Image} & \multicolumn{2}{c}{\bf Video} \\
& \bf w/o DP &\bf w/ DP  & \bf w/o DP &\bf w/ DP 
\\ \midrule
Full Fine-tuning&92.09&77.61 \textcolor{red}{(-14.48)}& 94.22&86.34 \textcolor{red}{(-7.88)}\\
Head-tuning&72.55&62.20 \textcolor{red}{(-10.35)}&88.57&83.09 \textcolor{red}{(-5.48)}\\
\midrule
FedPEFT-Bias &91.02& \textbf{84.98} \textcolor{red}{(-6.04)}&92.34&86.61 \textcolor{red}{(-5.73)}\\
FedPEFT-Adapter&88.05&79.05 \textcolor{red}{(-9.00)}&92.82&85.57 \textcolor{red}{(-7.25}\\
FedPEFT-Prompt &89.90&78.35 \textcolor{red}{(-11.55)}&93.82&\textbf{87.64} \textcolor{red}{(-6.18)}\\
\bottomrule
\end{tabular}%
}
\end{center}
\vspace{-15pt}
\end{table}
In this section, we further investigate our third research question \textbf{(RQ3)} in two critical FL scenarios evaluated by  CIFAR-100 and UCF-101.

\noindent\textbf{Differential Privacy.}
A fundamental property of federated learning is privacy protection. However, various works~\cite{huang_evaluating_2021, hatamizadeh_gradvit_2022} have demonstrated how the client data can be reconstructed from the raw gradient updates received by the server in some scenarios. To protect client data privacy from such attacks, differential privacy (DP) \cite{ kairouz_advances_2021,chamikara_local_2020,dwork_differential_2008, dwork_algorithmic_2014} has become standard practice.
Therefore, we first study FedPEFT and other baselines under DP.

To integrate DP, we apply a Gaussian mechanism within the local optimization of each iteration \cite{dwork_algorithmic_2014} with $\varepsilon=5$ and $\delta=0.001$. We maintain the remaining FL settings as described in Section~\ref{section:setting}, and show the results in Table~\ref{tab:dp}.
Interestingly, when comparing all methods, full fine-tuning experiences the sharpest drop with DP. This causes its accuracy to fall lower than all the FedPEFT prototypes. To understand this effect, we note that DP applies noise to all trainable parameter gradients. Full fine-tuning, therefore, requires such noise on all model parameters, resulting in a more pronounced negative effect on final performance. On the other hand, the other fine-tuning methods maintain some part of the backbone frozen and have significantly fewer trainable parameters on which adding noise is necessary, limiting the performance drop. 
Overall, FedPEFT allows for stronger accuracy in DP-enabled federated systems than even full fine-tuning while still maintaining extremely low communication needs.
%

\begin{table}[t]
\vspace{-3mm}
\caption{\small\textbf{Robustness analysis for data scarcity.} $K$ indicates the total sample number of all clients. }
\label{tab:few}
\vspace{-5pt}
\centering

\resizebox{0.45\textwidth}{!}{%
\begin{tabular}{l|cc|cc|cc}
\toprule
\multirow{2}{*}{\bf Method}& \multicolumn{2}{c|}{\bf $K=1000$}&\multicolumn{2}{c|}{\bf $K=1500$}& \multicolumn{2}{c}{\bf $K=2000$}\\ 
& Image & Video & Image & Video & Image & Video\\
\midrule
Full Fine-tuning&66.52&87.50&67.47&88.34&77.67&90.54 \\
Head-tuning & 52.13& 83.46 & 56.52&85.56 & 60.15&86.76\\
\midrule
FedPEFT-Bias & \textbf{76.40}&85.34 & \textbf{81.14}&87.36 & \textbf{83.83}&88.85 \\
FedPEFT-Adapter & 71.34&86.17 & 76.91&88.12 & 79.22&90.28 \\
FedPEFT-Prompt & 63.77&\textbf{87.46} & 71.94&\textbf{88.22}  & 76.89& \textbf{90.45}\\
\bottomrule
\end{tabular}%
}
\vspace{-10pt}
\end{table}
\noindent\textbf{Data Scarcity.} 
We explore another common yet challenging robustness condition in FL; that is when very little data is available on individual clients. 
Such data scarcity scenarios are even a tricky problem in centralized training. 
Fewer training data will incur damage to the pre-trained representation due to overfitting. 
In our evaluation for FL, we reduce the total sample number $K$ to $1000$, $1500$, and $2000$. 
As shown in Table~\ref{tab:few}, we find that FedPEFT outperforms full fine-tuning and head-tuning under such low-data scenarios, further revealing its capability to appropriately adapt pre-trained representations to the FL task at hand.

\rev{For the inter-prototype comparison, FedPEFT-Bias and FedPEFT-Prompt remain leading the performance in image and video domains, consistent with the conclusion in common scenarios, showing their robustness.}

\subsection{Insights and Takeaways}
\begin{colorsection}{purple}
Our research findings contribute significant insights to leverage parameter-efficient fine-tuning in federated learning, aiming at reducing communication costs. 

\begin{itemize}
\item FedPEFT stands out under stringent communication budgets by offering significant advantages over traditional approaches, such as reducing the number of participating clients, utilizing smaller models, or solely training the classification head. Remarkably, the total communication overhead for $50$ rounds in the FedPEFT framework is less than that of a single round in a conventional FL setting.
\item In scenarios without communication limitations, FedPEFT can achieve server accuracies exceeding $95\%$ while only requiring less than $0.3\%$ of the parameters to be trainable.
\item The continuum of fine-tuning freedom, ranging from Full fine-tuning through FedPEFT-Bias, FedPEFT-Adapter, FedPEFT-Prompt, to Head-tuning, varies across different scenarios. Typically, FedPEFT-Bias and FedPEFT-Prompt emerge as the top performers in image and video processing tasks. Notably, FedPEFT-Prompt demonstrates superior adaptability when bridging larger domain gaps between pre-trained models and downstream tasks.
\item The effectiveness of FedPEFT is further illustrated in real-world applications characterized by stringent privacy requirements or data scarcity. Even under such conditions, FedPEFT-Bias and FedPEFT-Prompt maintain exceptional performance across image and video domains.
\end{itemize}
\end{colorsection}

%% file: 5_discussion.tex
\pdfoutput=1


\section{Conclusion}
In this paper, we introduce FedPEFT, a new federated learning framework leveraging strong pre-trained models and massively reducing communication costs. 
We integrate three effective prototypes within the FedPEFT framework: Bias, Adapter, and Prompt. 
With a thorough empirical study, we then evaluate FedFEFT and other baselines in three key areas: communication, capability, and robustness. We find FedPEFT to be a promising approach for practical FL systems, capable of handling many of the harsh conditions in FL while alleviating the critical communication bottleneck.
As a general framework, FedPEFT can also be leveraged with other PEFT methods and in application domains other than computer vision. 
We hope this work can inspire new perspectives in federated learning through the combined innovation of strong pre-trained models and parameter-efficient fine-tuning methodologies.